\documentclass[letterpaper, 10 pt, conference]{ieeeconf}  

\IEEEoverridecommandlockouts                              
\usepackage{float} 
\usepackage{tabu}
\usepackage{textcomp}
\usepackage{mathtools}
\usepackage{hyperref}
\usepackage{amssymb}
\usepackage{bbm}
\usepackage{graphicx}
\usepackage{multicol}
\usepackage{multirow}
\usepackage[font=small,labelfont=bf]{caption}
\usepackage{subcaption}
\usepackage{algorithm}
\usepackage{bbm}
\usepackage{algorithmicx}
\let\labelindent\relax
\usepackage{enumitem,kantlipsum}
\usepackage{tablefootnote}
\usepackage{booktabs}
\usepackage{dsfont}
\usepackage[dvipsnames, svgnames]{xcolor} 
\usepackage{gensymb}
\usepackage{wrapfig}
\usepackage{capt-of}
\usepackage{xcolor}

\usepackage{algpseudocode}
\usepackage[backend=biber,
            hyperref=true,
            url=false,
            isbn=false,
            doi=false,
            backref=false,
            style=ieee,
            natbib=true,
            citestyle=numeric-comp,
            sorting=none,
            block=none]{biblatex}
\renewcommand{\bibfont}{\small}
\newcommand{\acronym}{\textsc{GeT-USE}}
\newcommand{\tog}{TOG-Net}
\newcommand{\hookandgrasp}{\texttt{Hook\_and\_grasp}}
\newcommand{\sweeping}{\texttt{Sweeping}}
\newcommand{\decanting}{\texttt{Decanting}}

\newsavebox{\preliminaries}
\newsavebox{\toolbuilding}
\newsavebox{\toolbuildingq}
\newsavebox{\toolgrasping}
\newsavebox{\toolgraspingq}
\newsavebox{\toolgraspingr}
\newsavebox{\actionspace}
\newsavebox{\toolmanipulation}
\newsavebox{\toolmanipulationq}
\newsavebox{\toolmanipulationr}
\newsavebox{\tooluseobjective}
\newsavebox{\environment}
\newsavebox{\toolgraspingobjective}
\newsavebox{\tooluseobjectiveapproximated}
\newsavebox{\tooluseobjectiveapproximatedmelt}
\let\ACMmaketitle=\maketitle
\renewcommand{\maketitle}{\begingroup\let\footnote=\thanks 
\ACMmaketitle\endgroup}
\makeatletter
\usepackage{graphicx}
\let\@fnsymbol\@arabic
\makeatother
\newcommand{\theappendix}{\@Alph\c@section}
\usepackage{etoolbox}
\newcommand{\debugcounter}[1]{%
  \typeout{DEBUG: Counter #1 = \arabic{#1}}%
}
\newcommand{\methodname}{\textsc{GeT-USE}}

\pretocmd{\section}{\debugcounter{section}}{}{}
\pretocmd{\subsection}{\debugcounter{subsection}}{}{}
\allowdisplaybreaks
\title{\LARGE \bf \methodname{}: Learning Generalized Tool Usage for Bimanual Mobile Manipulation via Simulated Embodiment Extensions}

\author{Bohan Wu, Paul de La Sayette, Li Fei-Fei\textsuperscript{*}, and Roberto Mart\'{i}n-Mart\'{i}n\textsuperscript{*}
\thanks{* Equal Contribution. Authors are with Stanford University and the University of Texas at Austin, USA. \texttt{\{bohanwu, pauldls, feifeili\}@stanford.edu, robertomm@utexas.edu}}
}

\begin{document}
\maketitle
\begin{abstract}
The ability to use random objects as tools in a generalizable manner is a missing piece in robots' intelligence today to boost their versatility and problem-solving capabilities. State-of-the-art robotic tool usage methods focused on procedurally generating or crowd-sourcing datasets of \textit{tools for a task} to learn how to grasp and manipulate them for that task. However, these methods assume that only one object is provided and that it is possible, with the correct grasp, to perform the task; they are not capable of identifying, grasping, and using the best object for a task when many are available, especially when the optimal tool is absent. 
In this work, we propose $\acronym$, a two-step procedure that learns to perform real-robot generalized tool usage by learning first to extend the robot's embodiment in simulation and then transferring the learned strategies to real-robot visuomotor policies. 
Our key insight is that by exploring a robot's embodiment extensions (i.e., building new end-effectors) in simulation, the robot can identify the general tool geometries most beneficial for a task. 
This learned geometric knowledge can then be distilled to perform generalized tool usage tasks by selecting and using the best available real-world object as tool. 
On a real robot with 22 degrees of freedom (DOFs), GeT-USE outperforms state-of-the-art methods by 30-60\% success rates across three vision-based bimanual mobile manipulation tool-usage tasks. 
\end{abstract}

\begin{refsection}[references.bib]
\section{Introduction}
\label{s_intro}

Tools are objects that facilitate achieving a manipulation task~\cite{stoytchev2007robot}.
Conceptually, they can be considered additions to the agent's embodiment: they enlarge it, equip it with a large flat area to swipe, or a convex form to contain other materials.
Learning to select and use the right tool for a task are critical skills to overcome the limitations of an embodiment~\cite{nabeshima2006adaptive} and they explain the astonishing manipulation capabilities of humans and some animals~\cite{vaesen2012cognitive}.
But while humans have created and perfected ad-hoc tools to support specific tasks ---hammers, screwdrivers, sweepers, dustpans, etc.---, in their absence, they are still capable of ``making the best'' of what is around, selecting and using the available object best-suited for a task.
For example, if we are tasked with wiping a surface from small objects but no sweeper is at hand, we may pick up a nearby magazine, napkin, sponge, or even hammer and push the object into a dustpan.
This \textit{generalized tool usage} capability is a cornerstone of human manipulation's versatility and problem-solving abilities; we would like to endow robots with similar skills.

While many robotic solutions have focused on using a task-specific tool~\cite{stuckler2014adaptive,stuckler2016mobile}, generalized tool usage is a more complex and less explored endeavor. 
The agent cannot assume there is \textbf{a} tool to identify, but it should instead consider the feasibility of achieving the manipulation goal with every available object.
A straightforward factorization where the robotic agent reasons about each present object, and how it should be grasped and used for the task would lead to a vast space of reasoning possibilities~\cite{toussaint2021co, toussaint2018differentiable}.
We posit that the agent should rather consider the geometric and morphological properties of the available objects to guide the selection and be equipped with a general manipulation skill capable of using a broad set of objects as tools for the task.
While existing approaches have learned to effectively identify the best tool for a task without learning to solve it with vision-based policies~\cite{brawer2020causal}, or grasp and manipulate individual objects for tool usage tasks~\cite{fang2020learning}, they struggle to evaluate multiple arbitrary objects, rank their task suitability, identify the optimal choice, and control their use using arm(s) and base motion based on onboard sensor signals.
A generic, versatile framework is missing for robots to learn general tool shapes and how to manipulate them based on sensor data.

\begin{figure}[t]
\centering
\includegraphics[width=\linewidth]{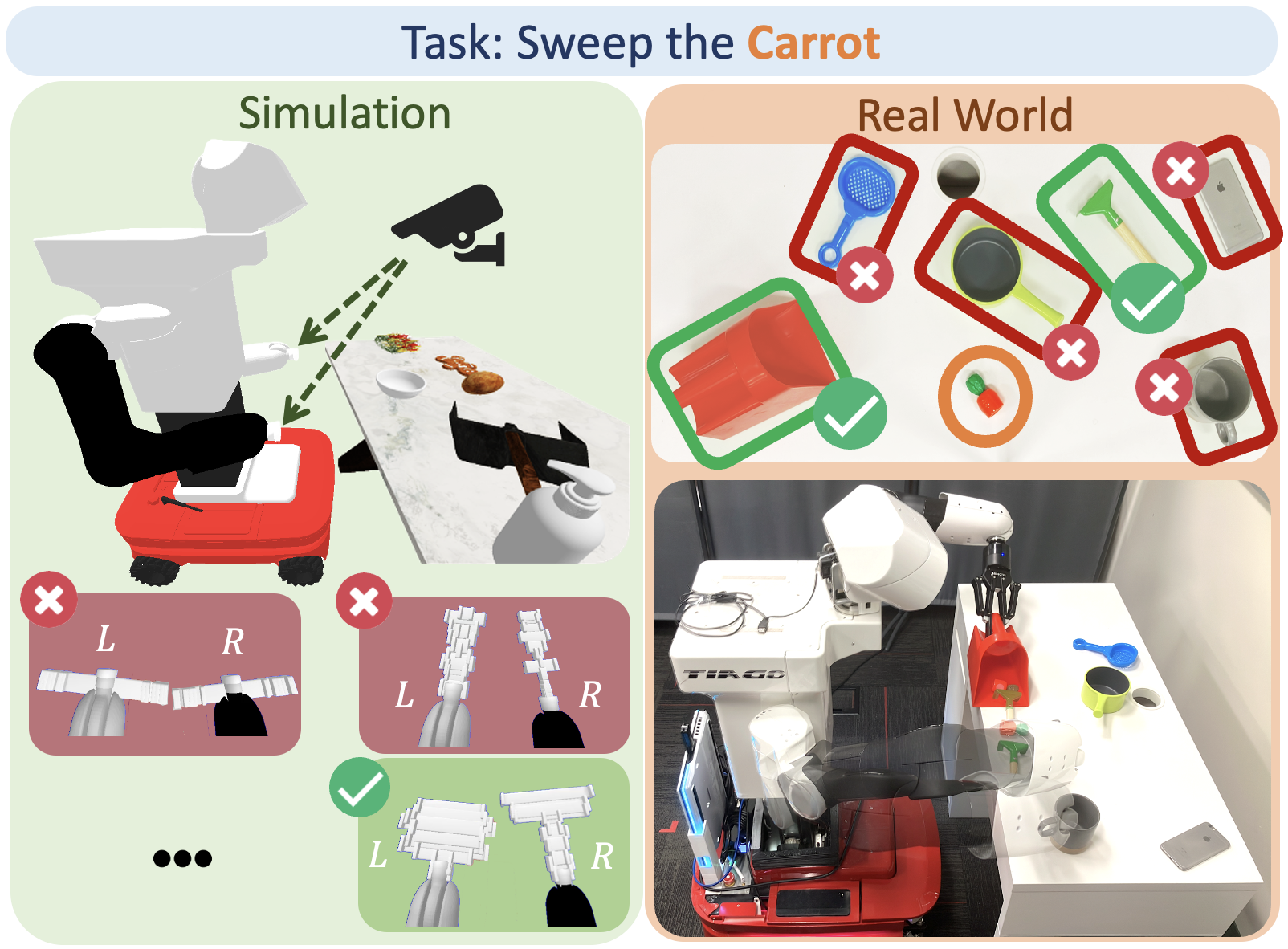}
\caption{\textbf{\methodname{}: \textbf{\underline{Ge}}neralized \textbf{\underline{T}}ool-\textbf{\underline{U}}sage via \textbf{\underline{S}}imulated \textbf{\underline{E}}mbodiment Extensions.} \textit{Left}: a TIAGo robot achieves a bimanual mobile manipulation task that requires using other objects as tools: ``Sweep the Carrot''. In simulation, \methodname{} explores different embodiment extensions by looking at (indicated by the ``camera'' icon) and building on top of its two wrists (``L'', ``R'') until it finds a suitable one. \methodname{} then transfers the successful strategy to the real world by learning vision-based modules to: 1) select the best available object (\textit{top right}), 2) grasp it (\textit{bottom right}), and 3) use it (\textit{bottom right}), all based on real depth images. This methodology allows the robot to learn generalized tool usage tasks in simulation that require bimanual mobile manipulation, and zero-shot transfer to the real-world.}
\label{pull}
\end{figure}

In this work, we propose a general two-step framework for robots to learn 1) the geometric properties and 2) the control strategies to achieve tasks using an available object as generalized tool, using one or two arms and moving the base when necessary.
We take inspiration from research in cognitive science that posits that when mastering tools, humans and animals consider the tool an embodiment extension of their own bodies~\cite{weser2021expertise,strauss1988tools}, and create a novel two-step procedure for learning to perform real-world generalized tool usage (Fig.~\ref{pull}): 
First, the robot learns in simulation to extend its own embodiment while controlling a bimanual mobile manipulation task using privileged information.
To learn embodiment extensions, we create a reinforcement learning approach with actions that control the exploration of the agent's morphologies.
Second, once a policy to create successful extensions is found, the robot grounds the geometric and morphological properties of the extensions and the control strategy in the visual domain. We use a teacher-student paradigm to generate 1) a visual affordance model to select the most suitable object in the surroundings, 2) a visuomotor policy to grasp it, and 3) another visuomotor policy to control the tool's usage for tool manipulation.
This enables a zero-shot sim2real transfer of the strategies learned in simulation into a physical bimanual mobile manipulator that uses only onboard sensor signals.
We call our method \textbf{\methodname{}} from \textbf{\underline{Ge}}neralized \textbf{\underline{T}}ool \textbf{\underline{U}}sage via \textbf{\underline{S}}imulated \textbf{\underline{E}}mbodiment Extensions.

We evaluated \methodname{} in three bimanual mobile manipulation tasks involving generalized tool usage: $\decanting$, $\sweeping$, and $\hookandgrasp$. $\acronym$ significantly outperforms existing state-of-the-art methods by 30-60\% success rates in these tasks based only on its onboard visual inputs by using generic objects as generalized tools.
This indicates that $\acronym$ is a versatile mechanism for robots to learn generalized tool usage by exploring extensions of their own body and transferring them to the real world, providing an efficient solution to learning to select and use objects as generalized tools in the real world.
\section{Related Work}
\label{s_rw}

\textbf{Learning tool usage in simulation.} 
Several prior works have attempted to learn to use tools for a single task in simulation~\cite{brown2011tool}. Two challenges need to be addressed: 1) how to generate virtual models of the tools, and 2) how to manipulate them. Several methods address the first challenge by procedurally generating tools based on engineered shape primitives~\cite{fang2020learning,bousmalis2018using,tobin2018domain, liu2023learning}, or crowd-sourcing object models from the internet~\cite{fang2020learning}. 
However, learning to select and make the best out of multiple available objects requires a much broader and more diverse training dataset of both good and bad tools, substantially increasing the burden of tool engineering and learning from crowd-sourcing and annotation (what tool categories are good/bad for \textit{hooking}?). 
In comparison, \methodname{} learns to detect, select, grasp, and manipulate the best object among the available ones in the real world, thanks to its simulation-trained tool-building policy that generates both suboptimal and good embodiment extensions to train tool selection, grasping, and manipulation models for generalized tool usage in the real world.

Several methods learned to use tools purely in simulation~\cite{stoytchev2005behavior,zhu2015understanding,gupta2021embodied}. However, they require privileged information to operate (e.g., the pose of the tool and objects), which is hard to obtain reliably in the real world. 
In contrast, \methodname{} relies on a policy with privileged information only to learn to build tools through embodiment extensions in simulation, but it then distills the information into vision-based policies that can operate directly from the robot's onboard sensor signals. 
Therefore, the final learned policies can perform real-world bimanual mobile manipulation tasks (Sec.~\ref{sec:experiments}). 

\begin{figure*}[t]
\centering
\includegraphics[width=\textwidth]{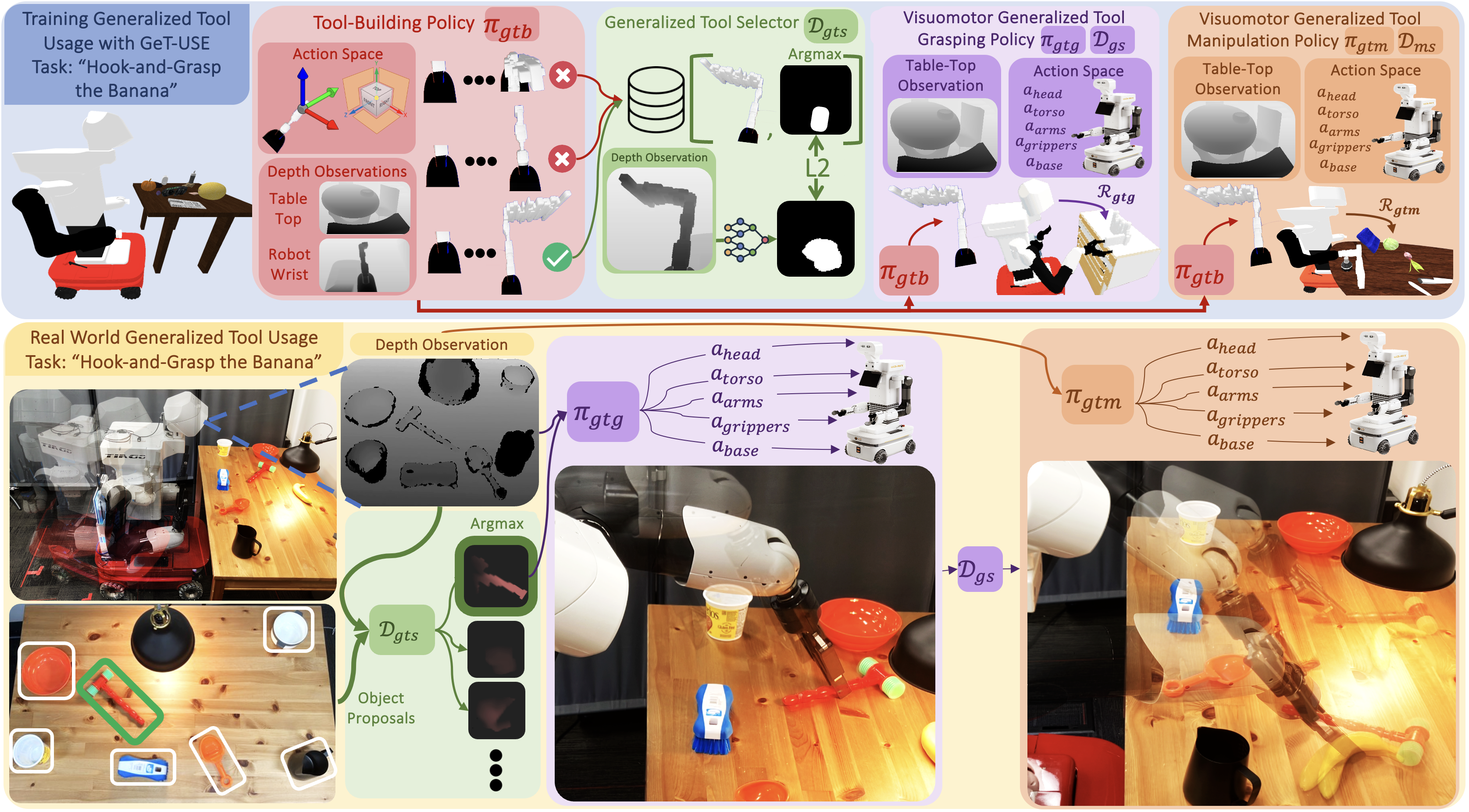}
\caption{\textbf{The \methodname{} Framework.} Below we explain training (\textit{\textcolor{blue}{top}}) and deploying in the real-world (\textit{\textcolor{yellow}{bottom}}) a generalized tool usage solution with \methodname{}. Training with \methodname{} (\textit{\textcolor{blue}{top}}) is a two-step procedure: In the first step, the agent is asked to solve a simulated version of the task (\textit{\textcolor{blue}{top, most left}}) and trains a generalized tool-building policy (\textit{\textcolor{red}{top, second-left}}), $\pi_\mathit{gtb}$, that explores by extending its own embodiment by appending elements (blocks) until it finds an extension that can be used to perform the task.
In the second step, the information of the tool-building policy is transferred to visual modules that can be used in the real world: i) a generalized tool selector (\textcolor{OliveGreen}{\textit{top, third-left}}), $\mathcal{D}_\mathit{gts}$, trained to predict the best grasping area (first generated blocks) of the generalized tools using successful (\textcolor{Green}{green tick}) and failed generated tools (\textcolor{red}{red cross}), ii) a visuomotor generalized tool grasping policy (\textit{\textcolor{Purple}{top, second-right}}), $\pi_\mathit{gtg}$, trained to grasp the tools created by $\pi_\mathit{gtb}$ using depth images as input and which success is detected automatically by a success detector, $\mathcal{D}_\mathit{gs}$, and iii) a visuomotor generalized tool manipulation policy (\textit{\textcolor{Purple}{top, second-right}}), $\pi_\mathit{gtm}$ that learns to output bimanual mobile manipulation commands to control the robot and achieve the task with tool generated by $\pi_\mathit{gtb}$. 
These modules are directly transferred and applied in the real world (\textcolor{yellow}{\textit{bottom}}): \methodname{} first captures a depth image of the table-top objects and generates object proposals. Each proposal is fed into \methodname{}'s tool selector, $\mathcal{D}_\mathit{gts}$ that selects the best object to use as generalized tool. \methodname{} then uses the tool-grasping and manipulation policies, $\pi_\mathit{gtg}$ and $\pi_\mathit{gtm}$, to grasp and manipulate the selected tool to achieve the task. This process allows \methodname{} to create solutions that solve tasks requiring generalized tool usage by leveraging simulation and transferring into visual modules for real world execution.
}
\label{fig:method}
\vspace{-5pt}
\end{figure*}

\textbf{Learning tool usage from real-world tools.} Earlier research efforts have tackled real-robot tool usage~\cite{hoffmann2014adaptive,sengul2013force} by gathering and training on extensive datasets comprising a wide variety of real-world tool objects and numerous real-robot trial-and-error~\cite{xie2019improvisation} or demonstration~\cite{qin2021rapidly} attempts at grasping and manipulating the tools. 
Such an extensive data-collection approach in the real world can avoid tackling the visual sim2real gap frequently encountered in simulation-based methods. However, scaling these methods to handle generalized tool usage involving suboptimal objects, and to control bimanual mobile manipulation from raw visual data would require significant computation resources and human labor. 
In contrast, \methodname{} uses depth observations to facilitate sim2real transfer and can scale successfully to bimanual mobile manipulation generalized tool usage tasks in the real world using large-scale simulation.

\textbf{Applying vision and foundation models to tool usage tasks}. Recent works have also used large language models~\cite{xu2023creative}, vision-language models~\cite{huang2024copa}, and object detectors~\cite{tee2018towards,huang2024copa} to learn tool usage. However, vision foundation models face limitations in precisely coordinating hand-eye movements, hindering their ability to perform tool usage tasks that require closed-loop visual feedback. In addition, tool usage success is fundamentally driven by the selected tool's local geometric features, in addition to global visual similarities to the ideal tool. Object detectors rank objects based on global visual similarities, and using them to select tools leads to sub-optimal downstream task performance, especially when the optimal tool is absent. In contrast, $\acronym$ learns to build tools using downstream task success as direct reinforcement learning rewards and performs bimanual mobile manipulation tasks that require fine-grained hand-eye coordination such as those in Sec.~\ref{sec:experiments}. 

\textbf{Planning for tool usage.} Prior methods have re-framed tool usage as a task planning problem and tried to plan how to grasp to perform downstream manipulation successfully~\cite{xu2021deep,agia2023stap,Toussaint2018DifferentiablePA}. However, they are either demonstrated solely in simulation or use planned primitives for tool-grasping and manipulation. Moreover, they scale poorly when they need to reason about the tool feasibility when many objects are available in the environment. 
Therefore, they cannot perform real-world bimanual mobile manipulation in the wild. 
\methodname{} overcomes these limitations by learning the main geometric properties of the successful embodiment extensions discovered in simulation and using them to select, grasp, and control the manipulation of the best-suited object in environments with multiple of them. 

To the best of our knowledge, no previous methods have attempted to learn to control all the degrees of freedom (arms, navigation base, torso, grippers, head) of a mobile manipulator for single-arm and bimanual generalized tool usage tasks using only onboard sensor signals.
\section{\small \textsc{GeT-USE}: Learning \underline{Ge}neralized \underline{T}ool-\underline{U}sage via \underline{S}imulated \underline{E}mbodiment Extensions}

\methodname{} is a two-step versatile procedure that allows robotic agents to learn generalized tool usage for different bimanual mobile manipulation tasks (see Fig.~\ref{fig:method}).
In the first step, \methodname{} learns to extend the robot's embodiment in simulation until it finds a modification that enables solving the tool usage task (Fig.~\ref{fig:method} \textit{top, second-left}, Sec.~\ref{ss:first}).
For this first step, the tool usage task is solved using a predefined strategy that exploits the simulator's privileged information.
After the successful embodiment extension is found, \methodname{} trains three vision-based modules necessary to transfer this information and perform generalized tool usage in the real world: a depth-map-based best generalized tool selector, a generalized visuomotor tool grasping policy, and a generalized visuomotor tool manipulation policy. 
This allows the robot to select the object in the environment that is best suited for a task to be used as generalized tool and perform the task with it in the real world, using only depth images from onboard sensors (Fig.~\ref{fig:method}, \textit{bottom}).

\methodname{} addresses several of the aforementioned steps as reinforcement learning problems which are modeled as Markov decision processes defined by the tuple $\mathcal{M} = \langle \mathcal{S}, \mathcal{O}, \mathcal{A}, \mathbb{T}, \mathcal{R}, \gamma, \rho_0, \mathcal{H}\rangle$. $\mathcal{S}$ is the state space (observable only in simulation), $\mathcal{O}$ is the space of observations. $\mathcal{A}$ is the action space. $\mathbb{T}$ is the dynamics model that governs the state transitions, $\mathbb{T}: \mathcal{S} \times \mathcal{A} \to \mathcal{S}$. $\mathcal{R}$ is a reward function. $\gamma \in [0, 1)$ is a discount factor. $\mathcal{H}$ is a finite horizon. $\rho_0$ is the distribution of initial states.
The goal of reinforcement learning is to train a policy, $\pi$, that takes optimal actions to maximize the expected future rewards with horizon $\mathcal{H}$.
Different steps in the \methodname{} procedure will be addressed as reinforcement learning problems with different actions, states, observation spaces, and rewards.
Below we describe these different subproblems in which \methodname{} factorizes the learning procedure for generalized tool usage task and specify those elements for the steps solved with reinforcement learning.

\begin{figure}[t]
\centering
\includegraphics[width=\linewidth]{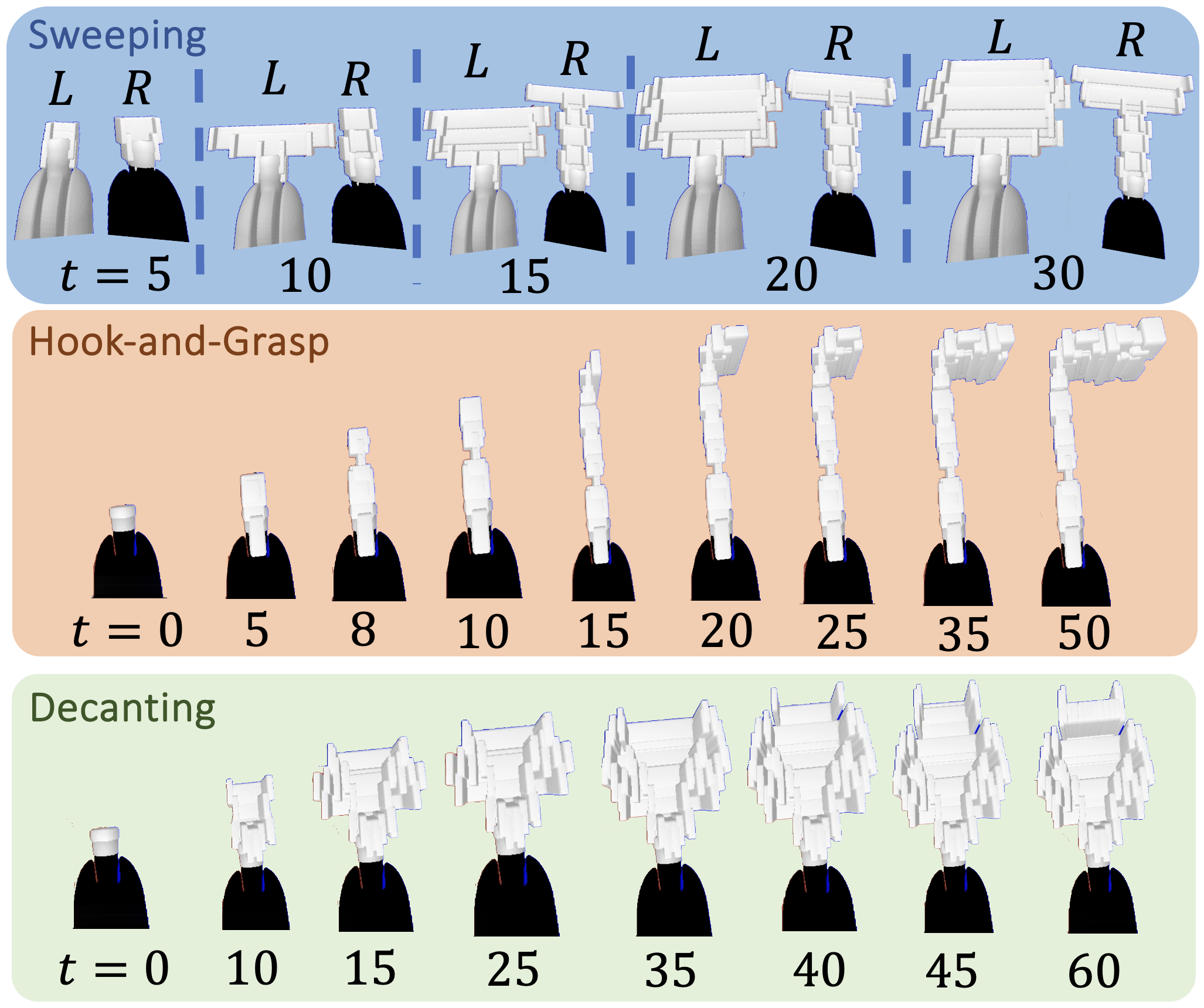}
\caption{Example rollouts of \methodname{}'s tool-building policy for $\sweeping$, $\hookandgrasp$, and $\decanting$ tasks. The bottom of each block details the tool geometry at timestep $t$ during the tool-building policy rollout. In \sweeping, \methodname{} builds one tool for each wrist with each timestep separated by a dashed line. The left (\textit{L}) wrist is marked gray; the right (\textit{R}) wrist is marked black. In \hookandgrasp~and~\decanting, the robot builds only one tool on either arm. As such, \methodname{} incrementally builds complex tools using small blocks one timestep at a time.}
\label{fig:process}
\end{figure}
\subsection{Learning a tool-building policy via embodiment extensions in simulation}
\label{ss:first}

In the first step, \methodname{} explores in simulation the space of possible extensions of its end-effector to achieve tool usage tasks (Fig.~\ref{fig:method}, \textit{top, second-left}).
This search is performed in simplified simulated versions of the tasks (in our experiments, $\decanting$, $\sweeping$, and $\hookandgrasp$, see Sec.~\ref{sec:experiments}).
To explore embodiments, \methodname{} trains first a tool-building policy, $\pi_\mathit{gtb}$, that incrementally generates and appends new elements to the robot's wrists before trying to solve the task. 
The elements appended are small cubes of variable size, max. 2~\text{cm}$^3$.
Using these elements, the policy can generate many helpful tools as they form multiple task-relevant shapes: hooks, sweepers, concave containers, etc.
Concretely, the tool-building policy receives as observations, $o \in\mathcal{O}_\mathit{gtb}$, two depth images from the simulator, one capturing the task objects (e.g., the object to hook, the object to sweep-clean, etc.), and the other capturing the wrist(s) of the robot where the tool(s) are being built.
The action space for the policy, $\mathcal{A}_\mathit{gtb}$, includes an $\mathbb{R}^3$ vector specifying the position of the new block relative to the previous block and an $\mathbb{R}^3$ vector specifying the dimension of the new block to append.
When the policy considers a suitable tool has been built, it terminates the episode, which triggers the automatic execution of a predefined manipulation strategy that uses privileged information from the simulator, such as the location of the task-relevant objects. 
The successes and failures of these predefined manipulation strategies provide rewards, $\mathcal{R}_\mathit{gtb}$, to train the tool-building policy. Fig.~\ref{fig:process} depicts an example rollout of the trained tool-building policy for each of the three tasks in our experiments (Sec.~\ref{sec:experiments}). 

Once \methodname{} has trained a tool-building policy in simulation, it uses it to train three vision-based modules that allow it to perform generalized tool usage in the real world.

\subsection{Learning to select generalized tools from depth images}
\label{ss:second}

As the next step, \methodname{} uses the trained tool-building tool (and the experiences of successful and failed tools) to train a visual generalized-tool selector module, $\mathcal{D}_\mathit{gts},$ that identifies the object in the environment best suited to be used as a tool for a task (Fig.~\ref{fig:method}, \textit{top, third-left}).
To do so, the final simulated depth image of the generated tool in each training episode of the tool-building policy is labeled with success or failure.
Depth images of a failed generated tool are associated with a binary image where all pixels are 0's.
Depth images of a successful generated tool are associated with a binary image where the ``handle'' of the tool (the first few blocks appended, the closest to the robot's wrists, which corresponds to the region to grasp) is marked as 1, and the rest of the image's pixels are marked as 0's (``binary image'' under ``Argmax'' in Fig.~\ref{fig:method}, \textit{top, third-left}).
\methodname{} then trains a tool-selector module that learns to output a binary image from an input depth image by using $l_2$-loss between the predicted binary images and ground-truth ones.
The activation in the output of the trained module thus acts as a likelihood estimator of the usefulness of an object, given an in-depth image of it.
In the real world. \methodname{} uses this tool-selector to select the object most promising to help as a generalized tool for the generalized tool usage task.

\begin{figure*}[t]
\centering
\includegraphics[width=\linewidth]{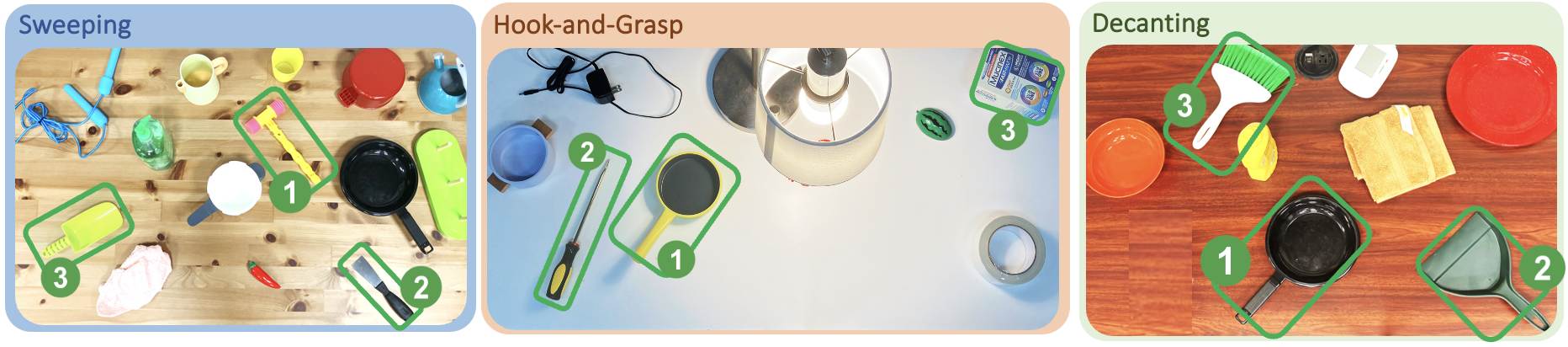}
\caption{\textbf{Example object preferences output by \methodname{}'s generalized tool selector module}. \methodname{}'s generalized tool selector ranks each object based on their suitability to serve as a sweeper in \sweeping~(\textit{left}), a hook in \hookandgrasp~(\textit{middle}), and a decanter in \decanting~(\textit{right}). The green ``1'', ``2'', and ``3'' in each block represent the tool selector's preference with ``1'' being the highest ranked. In this way, \methodname{} ``makes the best of what it has'', even when the ideal tool does not exist on the table.} 
\label{fig:selection}
\vspace{-5pt}
\end{figure*}

\subsection{Learning a visuomotor generalized tool-grasping policy}
\label{ss:third}

Selecting the object most promising to be used as a generalized tool is not sufficient to perform the task.
As the next step, \methodname{} trains a visuomotor tool-grasping policy, $\pi_\mathit{gtg}$ that uses depth images to control a grasping strategy on these objects (Fig.~\ref{fig:method}, \textit{top, fourth-left}).
To train this policy, \methodname{} calls at the beginning of the episode the tool-building policy, generates a suitable tool, detaches it from the simulated robot embodiment and places it randomly in the simulated scene.
The robot is then equipped with regular robot end-effectors (parallel Robotiq hands, see Sec.~\ref{sec:experiments}) and is trained to grasp the detached generated tool.
To that end, the tool grasping policy is given as input two depth images, one of the scene and one of the tool to grasp, and the robot's proprioception (joint angles), and outputs actions that include commands to move the base, torso, head, grippers, and arms of the bimanual mobile manipulator in joint space.
\methodname{} uses as dense reward the relative pose between the robot's wrists and the base of the generated tools, and a final binary reward if it achieves success in the task, i.e., if the generated tools are grasped such that their pose is the same as when they were generated as embodiment extensions.

\methodname{} also trains a grasp success detector (Fig.~\ref{fig:method}, \textit{top, fourth-left}), $\mathcal{D}_\mathit{gs}$, that uses the depth images to infer when the grasping step has been successfully performed by predicting, based on the images, when the final reward will be provided.
This module enables autonomous transition and execution of the \methodname{} visual modules on the real robot.

\subsection{Learning a visuomotor tool-manipulation policy}
\label{ss:fourth}
As the final step to be able to perform generalized tool usage in the real world, \methodname{} trains a visuomotor tool manipulation policy, $\pi_\mathit{gtm}$, that uses depth images to control the arm(s) to perform the task after the objects that serve as generalized tools have been grasped (Fig.~\ref{fig:method}, \textit{top right}).
To that end, \methodname{} uses the trained tool-building policy again to create useful embodiment extensions at the beginning of each episode.
Next, the tool manipulation policy receives a simulated depth image at each timestep, and controls the robot's base, torso, head, grippers, and arm(s) at each timestep to achieve the tool usage task with the created extension.
The policy is trained in simulation using success and failure of the task as a binary reward and auxiliary dense reward to get closer to the task-relevant objects. 

\methodname{} also trains a manipulation success detector (Fig.~\ref{fig:method}, \textit{top right}), $\mathcal{D}_\mathit{ms}$, that uses the depth images to infer when the tool usage task has been achieved by predicting, based on depth, when the final reward will be provided.

\section{Deploying \methodname{} Policies in Real-World}
\label{s:rw}
At test time, the robot is randomly placed in the real world with a set of objects arranged randomly on a table and needs to perform a task that requires using a tool (Fig.~\ref{fig:method} \textit{bottom left}).
To that end, the robot first uses an off-the-shelf 2D object detector to detect and crop depth-image patches of objects on the table.
The robot then uses \methodname{}'s visual tool selector on each of the patches of the objects and selects the one with the highest pixel value in the predicted binary image, which indicates the highest probability of including an object suited to be used as a generalized tool (Fig.~\ref{fig:method} \textit{bottom, second-left}). 
This depth image patch is then passed to the visuomotor tool grasping policy together with a depth image of the entire environment, which the policy uses to grasp the object in the right manner to be used as a generalized tool (Fig.~\ref{fig:method} \textit{bottom, third-left}).
Once the grasp success selector indicates the grasp is successful, the robot uses \methodname{}'s trained visuomotor tool-manipulation policy to perform the task, until its success detector indicates task success (Fig.~\ref{fig:method}, \textit{bottom right}). 
\begin{figure*}[t]
\centering
\includegraphics[width=\linewidth]{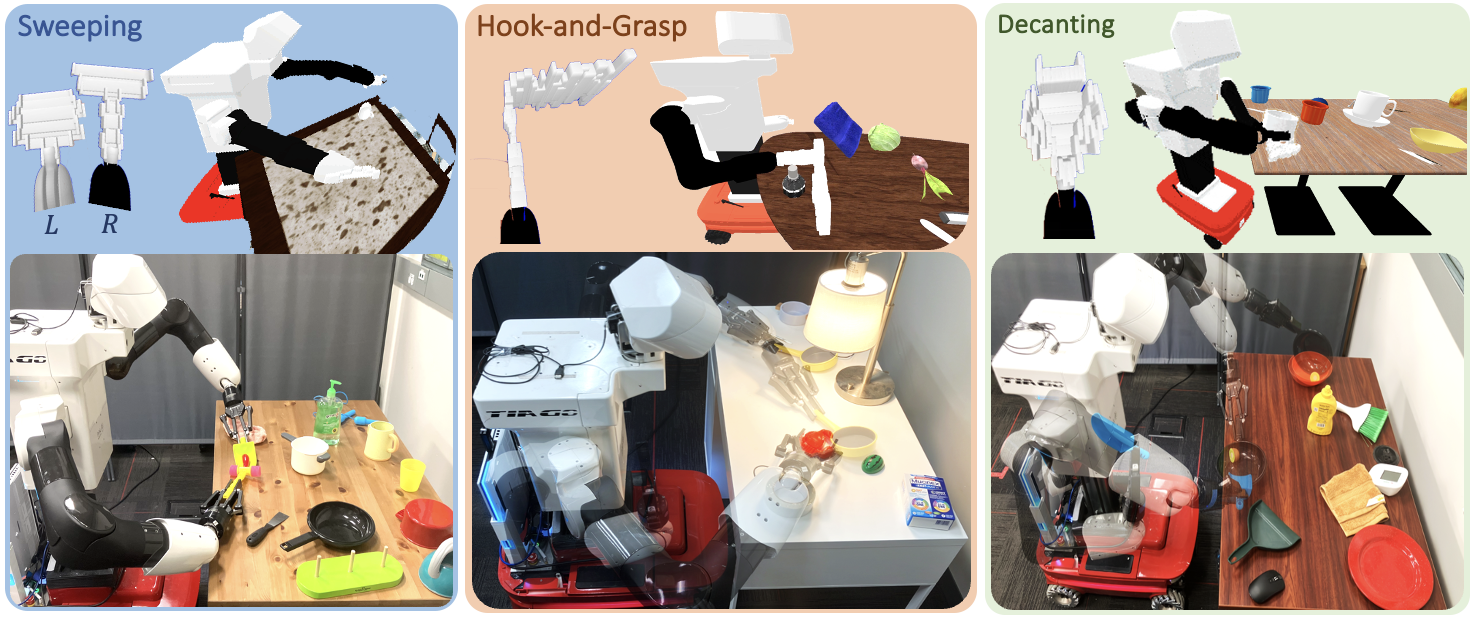}
\caption{\textbf{Simulated and Real-World Version of the Tasks.} \textit{Left to right.} Three tasks in our experiments: $\sweeping$, $\hookandgrasp$, and $\decanting$. \textit{Top}: example tool(s) generated by \methodname{}'s trained policy next to a simulation version of the tasks used for embodiment exploration and policy training of bimanual mobile manipulation tasks. 
\textit{Bottom}: example of the real-world version of the tasks. By learning from diverse tools and environments in simulation, \methodname{} successfully chooses the best tool in the real world, whether the ideal tool is present or not, and generalizes its policies to real-world objects, environments, and lighting conditions.}
\label{fig:experiments}
\vspace{-5pt}
\end{figure*}

\section{Experimental Evaluation}
\label{sec:experiments}

In our experiments, we use a TIAGo robot (Fig.~\ref{fig:experiments}, \textit{bottom}), a 22-DOF bimanual mobile manipulator with 3-DOF for the omnidirectional base, 7-DOF for each of the two arms, 1-DOF torso, 2-DOF head, and 1-DOF for each of the two Robotiq parallel-jaw grippers. The robot has an RGB-D camera mounted on its head. 
We use RGB-D images of $224 \times 224$ resolution at 3 Hz. 
In simulation, we only require simulated depth images.
In the real world, the RGB images are used by the 2D object detector that provides generalized tool candidates (see Sec.~\ref{ss:fourth}), while the depth image serves as input into \methodname{}'s tool selector, tool grasping and tool manipulation policies (Fig.~\ref{fig:method}, \textit{bottom}). 

A model of the same Tiago robot is used in simulation. We use iGibson~\cite{li2021igibson} to explore embodiment extensions and transfer the strategies to the real world. In iGibson, the robot is placed in front of a simulated version of each task and no tool objects and explores embodiments using \methodname{}. Simulation training is performed on 16 24GB GPUs with batch sizes of 128. 
\methodname{} uses 6-layer convolutional neural networks with three fully connected layers as the vision architectures of all modules. 

In simulation, we use a binary reward of 1 for the tool-building policy at the end of the tool-building episode if the predefined manipulation strategy (Sec.~\ref{ss:first}) succeeds at the tool usage task. 
We also leverage small, dense rewards to accelerate reinforcement learning adapted to each task (see details in the task explanations).
For the tool-grasping and tool-manipulation policy, we introduce a binary reward of 1 at the end of the training episode if tool-grasping or tool-manipulation succeeds. 
We also leverage a small reward negatively correlated with the distance between the end-effector and the built tool (for tool grasping) or the target object to manipulate (for tool manipulation). 

In the real world, the robot is placed in front of a table with multiple objects; one or more of them may be used successfully as generalized tools for the task, while others can be considered distractors and cannot lead to task execution. 
\methodname{} then controls the robot's bimanual mobile manipulation capabilities (base, torso, head, grippers, and arms) to grasp and achieve the task, using all DoFs simultaneously.
All experiments are conducted autonomously without human intervention from raw visual inputs with no QR codes. 
We use real-world objects (Fig.~\ref{fig:objects}), environments, and lighting conditions (Fig.~\ref{fig:experiments}) never seen during simulation training. 
In our experiments, we repeat each task 20 times with different initial locations and sampled objects.

\textbf{Tasks:} We evaluate \methodname{} in three challenging manipulation tasks that require generalized tool usage and the use of two arms and the base of the robot: \sweeping, \hookandgrasp, \decanting, which we describe below.

In \sweeping~(Fig.~\ref{fig:experiments}, \textit{left}), the goal is to sweep a target object randomly placed on the table. 
This is a common task in household scenarios when the robot needs to clean surfaces of small, hard-to-grasp objects or dirt.
In the real-world version of the task, we place 6 to 10 other objects randomly on the table, selected across 20+ object categories such as hammers and rakes (Fig.~\ref{fig:objects}), but \textbf{no actual sweepers are included.} 
The robot must choose two objects on the table and use one to push the object to sweep and the other to contain the object. 
Success is defined by whether the robot sweeps the target sweepable object up within 600 seconds. 
We then conduct a functional test by lifting both robot arms by 10 cm: the task is only successful if the object to sweep remains in the container without falling.
To train the tool-building policy in simulation, we add a small reward between 0 to 0.1 that is positively correlated with the Cartesian distance between the new block of the tool and the robot's tool-building wrist.

\begin{table*}[t]
\centering
\small
\caption{\textbf{Successful Trials} Out of 20 and Corresponding Success Rates for \methodname{}, State-of-the-Art Baselines and Ablations
} 
\resizebox{\linewidth}{!}{
\begin{tabular}{|c|c|c|c|c|c|c|c|c}
\toprule
 & \multirow{1}{*}{\methodname{}} & \tog-Internet & \tog-Proc & TOG-Net & \methodname{}-No-Ts & \methodname{}-Topdown-Grasp  & \methodname{}-Topdown-Manip \\
\midrule
\hookandgrasp & 16/20 & 9/20 & 11/20 & 0/20 & 4/20 & 0/20 & 2/20 \\ 
\sweeping & 11/20 & 6/20 & 2/20 & 0/20 & 2/20 & 0/20 & 0/20\\
\decanting & 10/20 & 4/20 & 1/20 & 0/20 & 1/20& 0/20& 0/20 \\\midrule
Total & 37/60 (61.7\%) & 19/60 (31.7\%) & 14/60 (23.3\%) & 0/60 (0\%) & 7/60 (11.7\%) & 0/60 (0\%)& 2/60 (3.3\%) \\
\bottomrule
\end{tabular}}
\label{tab:results}
\vspace{-15pt}
\end{table*} 

In \hookandgrasp~(Fig.~\ref{fig:experiments}, \textit{middle}), the goal is to retrieve and grasp a target object that is randomly placed on the far end of the table, outside of the reachable workspace of the robot. 
Additionally, the object is covered by a barrier that prevents the robot from successfully grasping it from above.
This is a common challenge for robots in unstructured environments, where objects may be out of its reach.
In the real-world version of the task, we place 3-5 other objects randomly on the table, selected among 20+ categories (Fig.~\ref{fig:objects}), including screwdrivers and pans, but \textbf{no actual hooks are included}. 
The robot must choose an object on the table to use as a generalized tool to hook the target object and grasp it with the second arm. 
The task is successful if the robot picks the target object and lifts it 10 cm above the table within 600 seconds. 
To train the tool-building policy in simulation, we add a small reward between 0 and 0.1 that is negatively correlated with the Cartesian distance between the last block of the tool and the target hookable object.

In \decanting~(Fig.~\ref{fig:experiments}, \textit{right}), the goal is to transfer a target object from a container into another object that serves as temporal container before decanting it into a final, third container.
This is a common scenario for robots that need to transfer small objects, particles, or even liquids from one place to another.
In the real-world version of the task, we place two containers randomly on the table: one larger, empty container (e.g., a bowl) that serves as the \textit{target container} and another small container (e.g., a spoon or scoop) that serves as the \textit{source container} and contains the small object to transfer. 
We place 3 to 5 additional objects randomly on the table, selected across 20+ categories, such as shovels and spoons (Fig.~\ref{fig:objects}). The robot should pick the small source container from the table, pour the contained object into the selected object, and then pour the small object again from the selected object into the target container. 
The task is successful if the robot pours the small object into the bowl via this two-step pouring process within 600 seconds. 
To train the tool-building policy in simulation, we add a small reward between 0 to 0.1 that is positively correlated with the distance between the new block of the tool and the wrist.

\textbf{State-of-the-Art Baselines:} We compare \methodname{} against TOG-Net~\cite{fang2020learning}, a state-of-the-art method for learning task-oriented tool grasping and manipulation.
We evaluate three variants of TOG-Net: 1) \textit{TOG-Net-Internet}, which learns from crowd-sourced Internet tool models and incorporates \methodname{}'s 6-DOF control and tool-selector. This variant corresponds to an ablated \methodname{} that crowd-sources Internet Tool Models instead of learning to build tools; 2) \textit{TOG-Net-Proc}, which learns instead from tools procedurally generated from simpler geometries such as sticks and incorporates \methodname{}'s 6-DOF control and tool-selector. This variant corresponds to an ablated \methodname{} that procedurally generates tools instead of learning to build tools; 3) \textit{TOG-Net}, which both crowd-sources Internet tools and procedurally generates tools without $\acronym$'s 6-DOF control or tool-selector.
We also compare against baselines of \methodname{}: 1) \textit{TOG-Net-No-Ts}, which selects \textit{randomly} an object to use as generalized tool, 2) \textit{TOG-Net-Topdown-Grasp}, which restricts grasping to only top-down rotation, and 3) \textit{TOG-Net-Topdown-Manip}, which restricts manipulation to only top-down rotation. These two last variants' action spaces were used in previous works such as \tog~\cite{fang2020learning}. 

In our experiments, we answer three main questions: 
\begin{figure}[t]
\centering
\includegraphics[width=\linewidth]{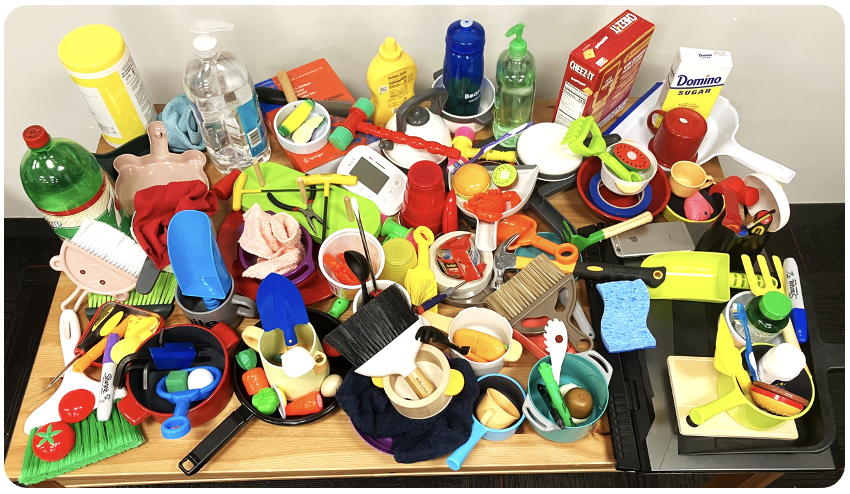}
\caption{All real-world objects used in our experiments for the three generalized tool usage tasks. They include useful and adversarial (useless) objects for each task. The ability to explore embodiment extensions in simulation enables \methodname{} to experiment with diverse tools and transfer vision-based modules that learned from them to the real-world to use diverse objects as tools.}
\label{fig:objects}
\end{figure}

\textbf{Q1: How does \methodname{} compare to state-of-the-art approaches that crowd-source or procedurally generate tools in real-robot generalized tool usage tasks?}
Table~\ref{tab:results} (columns 2-5) depicts the results of our comparison: \methodname{} outperforms all variants of TOG-Net by a significant margin.
\textit{TOG-Net} (Table~\ref{tab:results} \textit{column 5}) fails across all tasks since it cannot perform 6-DOF tool-grasping and manipulation or tool selection. 
\textit{TOG-Net-Internet} achieves some success thanks to the geometric similarities between Internet object models and real-world objects. However, due to limited models, \textit{TOG-Net-Internet} overfits ideal tool geometries, underfits bad tool geometries, and fails to complete most trials where no ideal tools exist. 
While \textit{TOG-Net-Proc} achieves success on \hookandgrasp~by procedurally generating hook-like objects using sticks, it fails mostly on \sweeping~and \decanting~due to the difficulty of hand-engineering sweeper-like and container-like tools from primitive sticks. 
This result validates \methodname{}'s design choice to learn to build tools because it enables the generation of a wide range of useful and useless tools, selection of good tools over bad tools, and ultimately, 30-60\% higher real-robot success rates (Fig.~\ref{fig:selection}) than all three versions of \tog. 

\textbf{Q2: How important is \methodname{}'s tool selector to generalized tool usage tasks?}
To answer this question, we compare \methodname{} (Table~\ref{tab:results}, \textit{column 2}) to \textit{TOG-Net-No-Ts} (Table~\ref{tab:results}, \textit{column 6}).
We observe a 50\% drop in performance for \textit{\methodname{}-No-Ts}. This indicates that learning to select the available object that can best serve as a tool for the task is a critical element to solve these tasks, and not all objects can be successfully used. 
\methodname{}'s tool selector is thus crucial to achieving generalized tool usage (answer to Q2).

\textbf{Q3: How important is it to control all 6-DOFs of the end-effector to grasp and manipulate generalized tools?}
To answer this question, we compare \methodname{} to \textit{\methodname{}-Topdown-Grasp} (Table~\ref{tab:results}, \textit{column 7}), which uses only top-down grasps, and ``\textit{\methodname{}-Topdown-Manip}'' (Table~\ref{tab:results}, \textit{column 8}), which uses only top-down manipulation. 
We observe that \textit{\methodname{}-Topdown-Grasping} and \textit{\methodname{}-Topdown-Manip} encounter 60\% and 57\% success rates degradation respectively compared to our full variant using all 6-DOF for grasping and manipulation. 
\textit{\methodname{}-Topdown-Grasping} fails in \hookandgrasp~because the target object is too far away from the top-down-grasped tool. In \sweeping, a sweeper tool object and a dustpan tool object grasped top-down make sweeping kinematically infeasible. In \decanting, a top-down-grasped container tool object makes pouring physically impossible. 
\textit{\methodname{}-Topdown-Manip} fails in \hookandgrasp~because the target hook-able object is so far away from the robot that it becomes unreachable by a top-down end-effector; in \sweeping, because the robot requires to move the end-effectors in 6-DoF to sweep and shovel successfully; and in \decanting, because both the first and second pouring motions require 6-DOF wrist rotation. 
Thus, 6-DOF grasping and manipulation are crucial for generalized tool usage.

\textbf{\methodname{} Failure Cases.} While achieving successes, \methodname{} fails in \sweeping~when the object the robot needs to sweep slides under the object the robot has selected to use as a generalized tool for sweeping. 
This happens when the tool manipulation policy rotates the generalized object for sweeping in a way that one part touches the table while the other is too far, creating a separation between the table surface and the edge to sweep with. For \hookandgrasp, \methodname{} fails when the object to hook and grasp is accidentally pushed with the generalized object to hook and ends outside of the reachable workspace. 
We hypothesize that the dynamics gap between simulation and real-world object manipulation is the main reason behind failure cases for both \sweeping~and \hookandgrasp. For \decanting, \methodname{} fails when the object in the generalized object used as temporal container misses the target container. 
This is due to poor execution of the real robot's motion controller, which creates a shaking vibration. 
This failure is caused by the robot's hardware limitations and we deem it independent of \methodname{}. 
Learning a better controller that can accurately control and alleviate the real robot vibrations would alleviate this error and increase success rates effectively, but would require a model of the disturbance in simulation, which we consider beyond the scope of our work. 
\section{Conclusion}
We proposed \methodname{}, a framework to learn real-robot bimanual mobile manipulation generalized tool usage tasks. 
Our method results from a two-step procedure that leverages first simulation to learn good tools for tasks as embodiment extensions, and then trains visuomotor modules to select, grasp and manipulate the best available object as tool in the real world.
Our method demonstrated success in three complex generalized tool usages.
Nevertheless, \methodname{} presents some limitations: \methodname{} assumes that the objects' movements can be accurately simulated and that the sim-to-real dynamics gap is small. 
While this assumption holds for most rigid-body objects, it does not hold for tasks with rich contacts, deformable objects such as fabric and clothes, or liquids such as water, and is thus left for future work. 
\methodname{} is also experimented only with parallel-jaw gripper grasping and manipulation, and adapting it to more dexterous tool usage tasks would require multi-fingered hands. 
This would enable \methodname{} to operate tools with articulation (e.g., scissors), but additional extensions may be required for these objects, which we left for future work. 

\section{Acknowledgment}
The authors sincerely thank Prof. Kenneth Salisbury and all the students in his laboratory for their support of this research.

\renewcommand*{\bibfont}{\footnotesize}
\printbibliography
\newpage
\end{refsection}
\end{document}